\PassOptionsToPackage{pdftex}{graphicx}
\documentclass{article}

\usepackage{arxiv}
\usepackage{multirow}
\usepackage{hhline}
\usepackage{xspace}
\usepackage{url}
\usepackage{amssymb}
\usepackage{scalefnt}
\usepackage{footnote}
\makesavenoteenv{table}
\makesavenoteenv{tabular}

\usepackage{times}
\usepackage{graphicx}
\usepackage{amsmath}
\usepackage{rotating}
\usepackage{setspace}

\usepackage{epstopdf}
\DeclareGraphicsExtensions{.jpg,.png,.pdf}
\usepackage[latin1]{inputenc}

\title{Binary and Multiclass Classifiers based on Multitaper Spectral Features for Epilepsy Detection}

\author{
	Jefferson Tales Oliva\\
	Academic Department of Informatics\\
	Federal University of Technology\\
	Pato Branco, Paraná, Brazil\\
	\texttt{jeffersonoliva@utfpr.edu.br} \\
	\And
	João Luís Garcia Rosa\\
	Department of Computer Science\\
	University of São Paulo\\
	São Carlos, São Paulo, Brazil \\
	\texttt{joaoluis@icmc.usp.br} \\
}

\begin{document}
	
	\maketitle

	\begin{abstract}
		Epilepsy is one of the most common neurological disorders that can be diagnosed through electroencephalogram (EEG), in which the following epileptic events can be observed: pre-ictal, ictal, post-ictal, and interictal. In this paper, we present a novel method for epilepsy detection into two differentiation contexts: binary and multiclass classification. For feature extraction,  a total of 105 measures were extracted from power spectrum, spectrogram, and bispectrogram. For classifier building, eight different machine learning algorithms were used. Our method was applied in a widely used EEG database. As a result, random forest and backpropagation based on multilayer perceptron algorithms reached the highest accuracy for binary (98.75\%) and multiclass (96.25\%) classification problems, respectively. Subsequently, the statistical tests did not find a model that would achieve a better performance than the other classifiers. In the evaluation based on confusion matrices, it was also not possible to identify a classifier that stands out in relation to other models for EEG classification. Even so, our results are promising and competitive with the findings in the literature.
	\end{abstract}

	\keywords{ Electroencephalogram \and epilepsy \and signal processing \and spectral features \and machine learning \and multiclass classification}

	\section{Introduction}
	
	According to the World Health Organization (WHO), about 50 million people have epilepsy, which sets it up among the most common neurological disorders~\cite{epilepsystatistics2019}. Of these patients, 80\% do not live in first world countries. In Brazil, for example, there are few studies on the incidence of this disease, for which most estimates were made for large cities (\textit{e.g.} São Paulo)~\cite{neto2005aspectos}. Based on the view that between four and ten per 1,000 people have epilepsy~\cite{epilepsystatistics2019}, in Brazil, between 840,588 and 2,101,471 inhabitants have this disorder\footnote{According to Brazilian Institute of Geography and Statistics (IBGE), the estimated Brazilian population in 2019 was 210,147,125 inhabitants~\cite{IBGE2020}.}.
	
	Epilepsy is a chronic disease denominated as a group of neurological disorders characterized by recurrent seizures, which are results of temporary electrical disturbances in brain cells~\cite{fisher2014ilae}. This disease is manifested in the following events~\cite{smith2005eeg,fisher2014can}:
	
	\begin{itemize}
		\item Pre-ictal: is a neurological activity prior to seizure.
		
		\item Ictal: is the event when the seizure occurs.
		
		\item Post-ictal: is the neurological event shortly after seizure.
		
		\item Interictal: is a neurological activity between seizures. Although this event is an evidence for epilepsy, it is considered normal (healthy patient).
	\end{itemize}

\begin{figure}[htb]
		\centering
		\includegraphics[width=120mm]{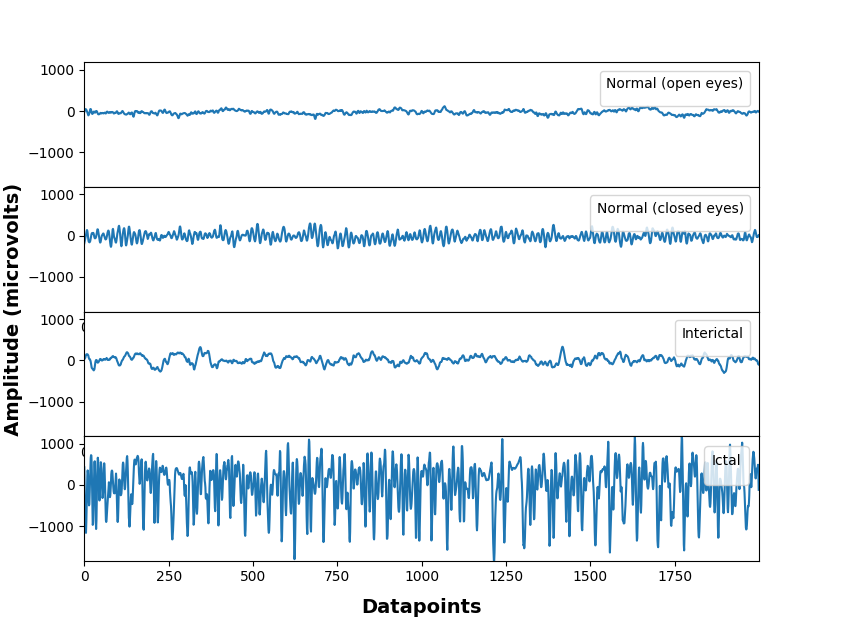}
		\caption{EEG segments examples.}
		\label{fig:heeg_segments}
	\end{figure}

	Epileptic events can be observed in electroencephalograms (EEG), which are the traces of the electrical brain activity over time~\cite{freeman2013a}. Figure~\ref{fig:heeg_segments} presents some EEG segment examples. In this image, a normal closed eye EEG segment contains a higher incidence of alpha waves (8--12 Hz) compared to the other normal signal. The interictal segment presents a higher incidence of slow waves. The ictal signal has higher peaks (negative and positive).
	
	EEG analysis is an essential task for neurological disorder diagnosis~\cite{shafi201725}. However, this analysis is considered complex because EEG can contain patterns difficult to be observed, even for experienced experts, which can express different opinions about an EEG pattern, making it necessary to use complementary examinations (\textit{e.g.} tomography) for an accurate diagnosis~\cite{oliva2017epileptic}.
	
	In this context, machine learning (ML) can provide support in diagnosis and decision-making processes. Thereby, this work aims to build binary and multiclass classifiers for epilepsy detection. 
	
	The organization of this paper is the following: Section~\ref{sec:related_work} presents some related works for epilepsy detection; Section~\ref{sec:features} exhibits signal processing methods used for feature extraction; Section~\ref{sec:classification} shows ML methods for classifier building; Section~\ref{sec:methods} describes the experimental settings applied in our study; Section~\ref{sec:results} presents results and discussion about our experimental evaluation; and Section~\ref{sec:conclusion} shows the main conclusions, contributions, and expected future works.

	\section{Related Works}
	\label{sec:related_work}
	
	In the literature, several methods were proposed for epilepsy detection in EEG segments using ML techniques. Some of these works (published from 2016) are summarized in this section.
	
	The authors in Ref. \cite{zeng2016automatic} presented a method to identify pre-ictal, interictal, and ictal epileptic events in EEG segments. For feature extraction, measures based on entropy and Hurst index~\cite{hurst1951long} were extracted. For classifier building, ML methods such as support vector machine (SVM)~\cite{cortes1995support}, linear discriminant analysis (LDA)~\cite{mclachlan2004discriminant}, nearest-neighbor (NN), and decision tree (DT)~\cite{alpaydin2014introduction} were computed. In an experimental evaluation, the LDA predictive model achieved the highest accuracy, which was 76.7\%.
	
	Ref. \cite{jaiswal2017local} proposed another method for epilepsy detection. For this, firstly, time-domain features, such as local neighbor descriptive pattern and one- dimensional local gradient pattern \cite{chatlani2010local}, were extracted. Then, artificial neural networks (ANN), DT, NN, and SVM methods were applied to build predictive models. In the evaluation, an ANN classifier reached the highest accuracy of 99.90\%.
	
	In another work, classifiers were built for differentiation between interictal and ictal EEG segments~\cite{wang2018epileptic}. To do so, discrete wavelet transform (DTW)~\cite{marchant2003time} and SVM methods were applied to extract features and build classifiers, respectively. As a result, the best performing model reached 99.9\% accuracy.
	
	Ref. \cite{goksu2018eeg} used DTW to build ANN classifiers for EEG segment differentiation between normal and ictal events. The classifier that had the highest performance reached 100\% accuracy.
	
	In Ref. \cite{subasi2019epileptic}, DTW and SVM were used to build classifiers aiming to differentiate EEG segments among normal and ictal classes. In an experimental evaluation, the best performing classifier achieved 99.75\% accuracy.
	
	Another study for epilepsy detection was conducted in Ref. \cite{ren2019classification}, aiming to differentiate normal, interictal, and ictal EEG segments. For feature extraction, autoregressive model~\cite{brockwell2016introduction}, DTW, and sample entropy~\cite{richman2000physiological} were computed. For predictive model building, extreme learning machine (ELM)~\cite{huang2006extreme} based on LDA was performed. As a result, the most accurate classifier reached 99.43\% accuracy.
	
	Ref. \cite{gao2020automatic} proposed a method for epilepsy detection based on time-domain features and ANN classification method. In an experimental evaluation, the predictive model reached 99.26\% accuracy.
	
	In Table~\ref{tab:related_work}, comparisons among our experimental evaluations and related works for epilepsy detection are presented. In these comparisons, the following criteria were used:
	
	\begin{itemize}
		\item Signal domain(s): how the EEG segments are represented for feature extraction. The signal representation can be in time domain (T), frequency domain (F), time-frequency domain (TF), and/or nonlinear approach (NL).
		
		\item Classification problem(s): related to the number of classes considered in the related work: binary (B), if two classes are considered; or/and multiclass (M), if the number of classes is above two.
		
		\item Classification method(s): ML techniques used to build classifiers.
		
		\item Amount of classifiers: number of predictive models built and evaluated.
		
		\item Evaluation measure(s): statistical approach(es) used for classifier evaluation.
		
		\item Statistical test(s): method(s) applied to verify the existence of statistical differences among predictive models.
	\end{itemize}

	\begin{sidewaystable}[!htp]
		\centering
		\scalefont{0.85}
		\renewcommand{\arraystretch}{1.10}
		\caption{Some summarized related works.}
		\label{tab:related_work}
		\begin{tabular}{|c|c|c|c|c|c|c|}
			\hline
			\textbf{Work} & \textbf{\begin{tabular}[c]{@{}c@{}}Signal\\ domain(s)\end{tabular}} & \multicolumn{1}{c|}{\textbf{\begin{tabular}[c]{@{}c@{}}Classification\\ problem(s)\end{tabular}}} & \textbf{\begin{tabular}[c]{@{}c@{}}Classification\\ method(s)\end{tabular}} & \textbf{\begin{tabular}[c]{@{}c@{}}Amount of\\ classifiers\end{tabular}} & \textbf{\begin{tabular}[c]{@{}c@{}}Evaluation\\ measure(s)\end{tabular}} & \textbf{\begin{tabular}[c]{@{}c@{}}Statistical\\ test(s)\end{tabular}} \\ \hline
			\cite{zeng2016automatic} & NL & M & \begin{tabular}[c]{@{}c@{}}DT, NN, LDA,\\ and SVM\end{tabular} & 20 & Accuracy (Acc) & Anova \\ \hline
			\cite{jaiswal2017local} & T & B and M & \begin{tabular}[c]{@{}c@{}}ANN, DT, NN,\\ and SVM\end{tabular} & 32 & \begin{tabular}[c]{@{}c@{}}Acc, Sensitivity (Sen),\\ and Specificity (Spe) \end{tabular} & --- \\ \hline
			
			\cite{shen2017ga} & TF & M & SVM & 1 & Acc & --- \\ \hline
			
			\cite{wang2018epileptic} & TF & B & SVM & 10 & \begin{tabular}[c]{@{}c@{}}Acc, Sen, Spe,\\ Positive predictive value (PPV),\\ and Negative predictive value (PPV)\end{tabular} & Anova \\ \hline
			\cite{sikdar2018epilepsy} & TF & M & SVM & 3 & \begin{tabular}[c]{@{}c@{}}Acc, PPV, Spe, Sen, and F1-score\end{tabular} & \begin{tabular}[c]{@{}c@{}} Anova and Tukey \end{tabular} \\ \hline
			
			\cite{yavuz2018epileptic} & TF & B and M & ANN & 16 & \begin{tabular}[c]{@{}c@{}}Acc, Sen, and Spe\end{tabular} & --- \\ \hline
			
			\cite{goksu2018eeg} & TF & B and M & ANN & 27 & \begin{tabular}[c]{@{}c@{}}Acc, Sen, Spe, F1-score, and\\Kappa\end{tabular} & --- \\ \hline
			\cite{subasi2019epileptic} & TF & B & SVM & 2 & \begin{tabular}[c]{@{}c@{}}Acc, Sen, and Spe\end{tabular} & --- \\ \hline
			\cite{ren2019classification} & T, TF, and NL & M & ELM & 6 & Acc & --- \\ \hline
			\cite{sayeed2019neuro} & T and TF & B & ANN and NN & 6 & \begin{tabular}[c]{@{}c@{}}Acc, Sen, PPV, and F1-score\end{tabular} & --- \\ \hline
			\cite{wu2019intelligent} & T and F & B & ANN and DT & 4 & Acc & --- \\ \hline
			
			\cite{gao2020automatic} & T & B & ANN & 1 & \begin{tabular}[c]{@{}c@{}}Acc, Sen, and Spe\end{tabular} & \begin{tabular}[c]{@{}c@{}}Shapiro-Wilk,\\Student's t, and\\ Mann-Whitney U\end{tabular} \\ \hline
			
			Ours & F, TF, and NL & B and M & \begin{tabular}[c]{@{}c@{}} LDA, ANN, QDA,\\SVM, NN, and DT \end{tabular} & 18 & \begin{tabular}[c]{@{}c@{}} Mean error and confusion matrix\\parameters \end{tabular} & \begin{tabular}[c]{@{}c@{}} Shapiro-Wilk,\\Friedman,\\and Nemenyi \end{tabular} \\ \hline
		\end{tabular}
	\end{sidewaystable}

	Also, in recent years, several deep learning applications were proposed for epilepsy identification, such as presented in references \cite{acharya2018deep,ullah2018automated,san2019classification,cho2020comparison}. However, we did not consider these methods in this paper due to the following reasons: we used a small EEG dataset in our experiment since, for the handling of deep learning techniques, the use of large databases is recommended for classifier building~\cite{kohli2016hierarchical}; and since the EEG segments used in our experiments can contain linearly separable patterns, for the best of our knowledge, conventional ML methods can build accurate predictive models.
	
	In ML applications for epilepsy detection, most related works are focused in two-class problems, such as normal \textit{vs.} interictal, normal \textit{vs.} ictal, interictal \textit{vs.} ictal, seizure-free \textit{vs.} ictal, etc. Like several other real problems, epilepsy detection can be considered a multiclass problem because, besides the existence of several types of this disorder, a seizure is divided into stages (pre-ictal, ictal, post-ictal, and interictal)~\cite{fisher2014can}. Also, normal EEG varies according to, for example, age and sleep stages~\cite{tatum2014ellen}. Even so, binary classification is important in epilepsy domain because differentiating between normal and abnormal EEG can be a difficult problem considering different variations, such as age or if an EEG segment was collected when the volunteer was with the eyes open or closed.
	
	In our work, we combined three different domain features (frequency, time-frequency, and nonlinear), which contributed to the generation of accurate predictive models. Likewise, multiclass classifiers were generated for differentiation of EEG segments among four classes (in the related works, up to 3 classes were considered).
	
	According to the best of our knowledge, the use of a statistical hypothesis test is still less used in the literature about ML applied in EEG segments. Finally, a different task in multiclass classifier evaluation is the use of confusion matrices (Section~\ref{subsec:performance_evaluation}), which is widely used in the assessment of binary predictive models. All these evaluation approaches were conducted in this work for both binary and multiclass classifiers.

	\section{Spectral Features}
	\label{sec:features}
	
	Spectral analysis is widely applied in EEG processing because it is a powerful tool that provides more informative data (patterns) in comparison to the signal represented in time domain~\cite{babadi2014review}. Signal analysis in spectral domains is commonly performed by variations of the Fourier transform, such as discrete Fourier transform (frequency domain), short-time Fourier transform (time-frequency domain), and Fourier transform of the third-order cumulant (nonlinear approach).
	
	The conventional Fourier transform has some limitations, since biased spectral components can be generated, making it unreliable~\cite{percival2010spectral}. So, multitaper~\cite{thomson1982spectrum} was proposed in order to minimize Fourier transform limitations. This method can be computed in a time domain signal $S$ with length $n$ using Equation~\ref{eq:multitaper:ps}, where $M_{i, k}$ is the $k$-th data taper, $\omega = 2 \pi k$ is the angular frequency, $j$ is the imaginary unit, and $e^{-i 2 \pi k \frac{j}{n}}$ is equivalent to the Euler's complex exponential function ($\cos x + j\sin x$)~\cite{moskowitz2002course}. In other words, part of this equation is a correlation between a signal and the Euler's function~\cite{oliva2018differentiation,oliva2018:phd}.
	
	\begin{equation}
	\label{eq:multitaper:ps}
	X_{k} = \sum\limits_{i = 1}^{n} M_{i, k} \text{ } S_{i} \text{ } e^{-j \omega \frac{i}{n}}
	\end{equation}

	In Equation~\ref{eq:multitaper:ps}, data tapers are used to reduce broadband and narrow-band in $X$~\cite{babadi2014review}. Data tapers are generated through Slepian sequences, which are originated from spectral concentration problem~\cite{slepian1978prolate}.
	
	Multitaper can transform a time domain signal into other representations from different domains, power spectrum\footnote{Frequency domain.} (PS), spectrogram\footnote{Time-frequency domain.} (SG), and bispectrogram\footnote{Nonlinear approach.} (BS)~\cite{oliva2018:phd,oliva2019epilepsy}.

	\subsection{Power spectrum}
	
	PS can be produced by absolute square of components generated through Equation~\ref{eq:multitaper:ps}, \textit{i.e.}, $P_k = |X_k|^2$. For EEG analysis, PS features are often extracted by power bands, such as delta (1--4 Hz), theta (4--8 Hz), alpha (8--12 Hz), beta (12--30 Hz), and gamma (30--60 Hz)~\cite{bacsar1980comparative,freeman2013a,li2016signal}.
	
\begin{figure}[htb]
		\centering
		\includegraphics[width=120mm]{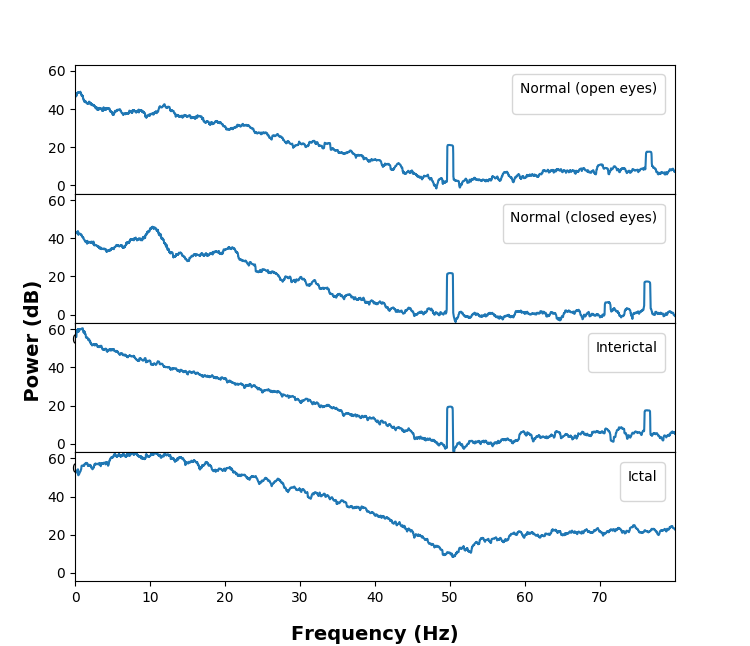}
		\caption{Power spectrum examples.}
		\label{fig:ps}
	\end{figure}

	In Figure~\ref{fig:ps}, PS examples, in their simplified views (converted to decibels (dB)\footnote{The $k$-th PS component (power) was computed by $10 * \log_{10}(P_k)$.}), extracted from different EEG segments, are illustrated. In this figure, ictal PS shows the highest incidence of theta, alpha, beta, and gamma waves. Interictal PS presents a higher incidence of delta waves. Regarding normal spectra, the highlight of this experiment is the higher incidence of alpha waves in closed eyes signal. Finally, peak oscillations at around 50 Hz (visible in all PS examples, except the ictal) are artifacts generally caused by power line pickup in electrodes~\cite{ferree2001scalp}. 
	
	For each power band, several features can be measured, \textit{e.g.} average, standard deviation, peak value, peak frequency, power ratio, centroid, kurtosis, skewness, first-order moment, second-order moment, Shannon entropy, root mean square, crest factor, flatness, and coefficient of variation~\cite{book:fredman:2011,oliva2018:phd}. Thereby, considering these 15 measures extracted by five power bands, a total of 75 (15 * 5) features can be extracted.

	\subsection{Spectrogram}
	
	By adapting Equation~\ref{eq:multitaper:ps}, multitaper can also generate SG. In this adaptation (Equation~\ref{eq:multitaper:sg}), time ($t$) must be considered. Thus, in comparison with short-time Fourier transform (STFT), which uses a window function to traverse signals, the adapted multitaper employs the data tapers as sliding windows~\cite{oliva2018:phd}.
	
	\begin{equation}
	\label{eq:multitaper:sg}
	SG_{t, k} = \left|\sum\limits_{i = t}^{t + l - 1} S_{i} M_{i - t + 1, k} e^{-j \omega \frac{i}{n}}\right|^2
	\end{equation}

\begin{figure}[htb]
		\centering
		\includegraphics[width=90mm]{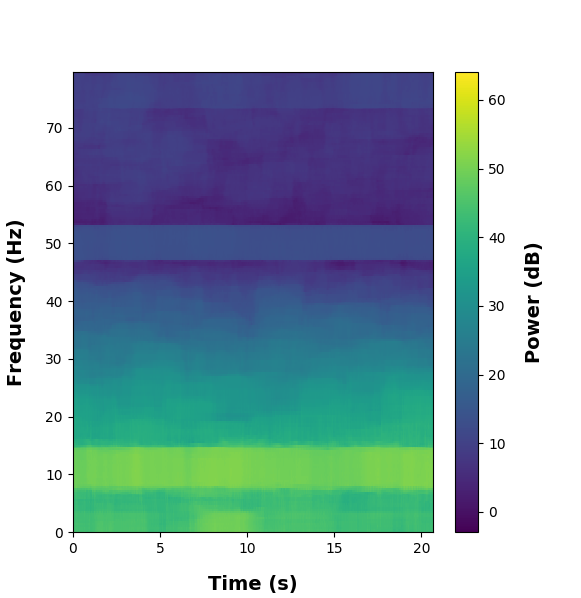}
		\caption{Normal spectrogram example.}
		\label{fig:healthy_sg}
	\end{figure}

\begin{figure}[htb]
		\centering
		\includegraphics[width=90mm]{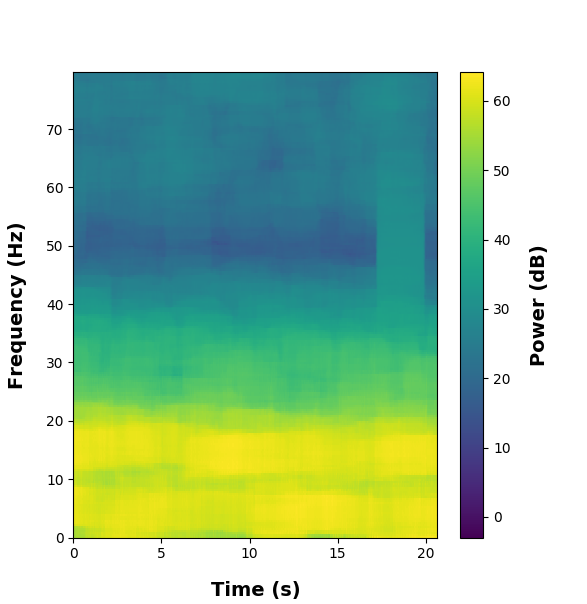}
		\caption{Abnormal spectrogram example.}
		\label{fig:ictal_sg}
	\end{figure}

	Figures \ref{fig:healthy_sg} and \ref{fig:ictal_sg} present SG examples generated from normal and abnormal EEG segments. The normal SG example was generated from a signal collected from a healthy volunteer with closed eyes. In this spectrogram, alpha waves are highlighted. In the abnormal SG example, which was created from an ictal EEG, contains higher intensity of spectral components above 20 Hz. More SG examples can be found in Ref. \cite{oliva2018:phd}.
	
	As in PS, SG features are also commonly computed by power bands (delta, theta, alpha, beta, gamma) in EEG analysis. For this, measures can be extracted for each power band, such as Shannon entropy, Renyi entropy, centroid, band energy, and bandwidth~\cite{oliva2018:phd,oliva2019classification}. Thus, a total of 25\footnote{Five measures were computed for each of the five bands.} features can be obtained for these measures.

	\subsection{Bispectrogram}
	
	The components generated by Equation~\ref{eq:multitaper:ps} can be used to create a BG, which is composed of third-order moment components computed through Equation~\ref{eq:hos}, where $f_{1}$ and $f_{2}$ are the analyzed frequencies and $X*$ is conjugate of the $X_{f_{1} + f_{2}}$ component. As presented in Equation~\ref{eq:multitaper:ps}, Each BG component is generated through the product of three PS components~\cite{acharya2013automated}.
	
	\begin{equation}
	\label{eq:hos}
	BG_{f_{1}, f_{2}} = |X_{f_{1}} X_{f_{2}} X^{*}_{f_{1} + f_{2}}|^{2}
	\end{equation}

	In the third-order spectral analysis, the dependence among $X_{f_{1}}$, $X_{f_{2}}$, and $X_{f_{1} + f_{2}}$ is verified. In this sense, the higher the value of $BG_{f_{1}, f_{2}}$, the higher is the dependence among the analyzed components~\cite{wong2009introduction}.

\begin{figure}[htb]
		\centering
		\includegraphics[width=90mm]{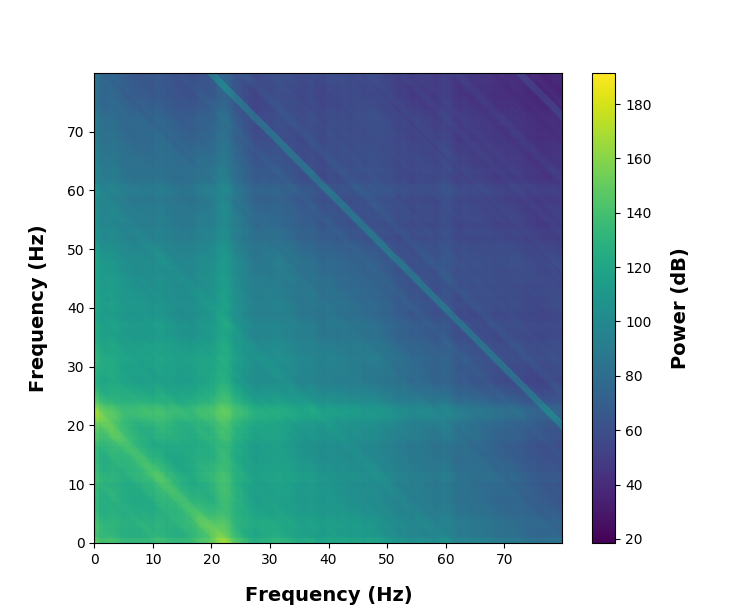}
		\caption{Normal bispectrogram example.}
		\label{fig:healthy_bg}
	\end{figure}

\begin{figure}[htb]
		\centering
		\includegraphics[width=90mm]{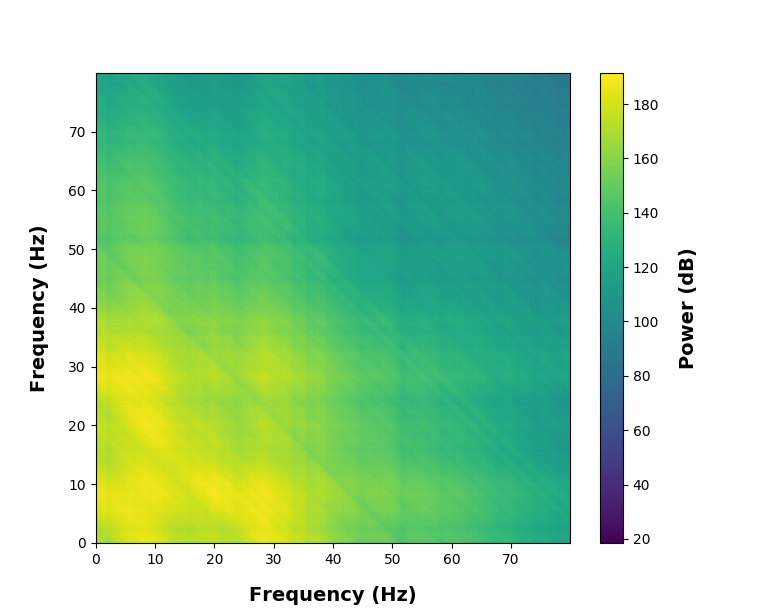}
		\caption{Normal bispectrogram example.}
		\label{fig:ictal_bg}
	\end{figure}

	Figures \ref{fig:healthy_bg} and \ref{fig:ictal_bg} present BG examples generated from normal (open eyes) and abnormal (ictal) EEG segments. In the healthy BG, the highest magnitude values are concentrated between 0 and 20 Hz, evidencing dependence among frequency components delta, theta, alpha, and part of the beta (12 to 20 Hz). In Figure~\ref{fig:ictal_bg}, all magnitude values were higher than those measured for the normal BG example, showing higher dependence between frequency components of the analyzed ictal EEG. Also, more BG examples can be found in Ref. \cite{oliva2018:phd}.
	
	In the feature extraction, only non-redundant area is considered in BG processing. This area is delimited by $0 \leq k_{2} \leq k_{1}$ and $k_{1} + k_{2} < n$~\cite{collis1998higher}. In this context, the following measures can be computed from a BG: mean magnitude, normalized entropy, normalized quadratic entropy, weighted center X, and weighted center Y~\cite{zhou2008classifying}.

	\section{Classifier Building}
	\label{sec:classification}
	
	Classifiers can be built by using sets of features as input of machine learning (ML) algorithms. For this, several ML approaches were proposed to solve classification problems, for example, binary and multiclass types, which are more predominant in the literature~\cite{book:Grigorev2017,manierre2018successful}.

	\subsection{Binary classification}
	\label{subsec:bin}
	
	Binary classification is the most common problem type in learning applications, in which only two classes (\textit{e.g.} normal and abnormal) are considered for predictive model building~\cite{book:Grigorev2017,manierre2018successful}.
	
	In this sense, ML methods from different paradigms were commonly applied in binary problems, of which some examples are presented below:
	
	\begin{itemize}
		\item Artificial neural network (ANN): is a learning method used to build connectionist classifiers based on the simplified view of the neuron. An ANN is composed of interconnected artificial neurons, commonly organized into layers~\cite{haykin2009neural}.
		
		\item Decision tree (DT): builds a hierarchically organized model by definition of rules, which are used to predict classes of input data. Thereby, the DT structure is represented by leaf and decision nodes, whose possible paths define the rules~\cite{quinlan1987simplifying}.  
		
		\item Linear discriminant analysis (LDA): is a technique used both for dimensionality reduction and classifier building by using Fisher's linear discriminant rule. In an ML application, LDA performs a linear transformation of the feature space, in which the pattern separation is maximized by classes~\cite{balakrishnama1998linear,mclachlan2004discriminant}.
		
		\item Nearest-neighbors (NN): is a lazy learning method whose classification is performed by measuring the similarity among input data and training examples (previously labeled) through distance measures, such as Euclidean distance~\cite{Mitchell:1997}.
		
		\item Support vector machine (SVM): is a learning method that maps patterns from a training set to an n-dimensional space through hyperplanes maximally separated. Each hyperplane contains margins used to separate the input data by class into a feature space~\cite{cortes1995support,suthaharan2016support}.
	\end{itemize}

	\subsection{Multiclass classification}
	\label{subsec:mul_classification}
	
	Most of the ML algorithms were developed only for binary problems. However, multiclass classification is required in large part of real applications~\cite{flach2012machine}.
	
	In this scenario, multiclass problems can be generalized by decomposition methods, which aim to divide them into binary subproblems, whose combination generates multiclass classification~\cite{lorena2011}. A multiclass decomposition method is applied in two steps:
	
	\begin{enumerate}
		\item Decomposition: a set of binary subproblems is generated from a multiclass problem. These subproblems can be represented by an $r \times c$ code matrix, where $r$ is the number of classes related to the multiclass problem and $c$ is the amount of binary subproblems. In a code matrix, values such as $+1$ (positive class), $-1$ (negative class), or $0$ (omitted class) can be instanced as element. For each code matrix column, a binary classifier is built considering only classes tagged with $-1$ or $+1$ labels. Thereby, $\frac{1}{2} * (3^{l} + 1) - 2^{l}$ binary classifiers can be built from a code matrix~\cite{mayoraz1997decomposition}. From the decomposition techniques presented in the literature, the all-against-all (AAA)~\cite{hastie1998classification} has the advantage of maintaining the original balance of training examples by class. The AAA approach generates a $\left(\frac{r * (r - 1)}{2}\right) \times c$ code matrix, \textit{i.e.}, all possible binary combinations are considered for a multiclass problem. In each subproblem, conventional ML methods (such as those presented in Section~\ref{subsec:bin}) can be performed\cite{oliva2019classification}.
		
		\item Reconstruction: a multiclass predictive model is given by an array of binary classifiers, whose output combinations determine the class of new examples. For this, the similarity between the output array and each code matrix row is computed by a distance measure (\textit{e.g.} Euclidean). The class of a new example is given by the code matrix row in which the lowest distance value was measured~\cite{passerini2004new}.
	\end{enumerate}

	\section{Experimental Evaluation}
	\label{sec:methods}
	
	\subsection{EEG Dataset and Feature Extraction}
	\label{subsec:eeg}
	
	To facilitate replication of our results, the publicly available EEG Epileptologie Bonn database\footnote{\url{http://epileptologie-bonn.de/cms/front_content.php?idcat=193&lang=3}}~\cite{andrzejak2001indications} was used. Another advantage of this database is the variety of type of EEG segments, which are divided into following sets:
	
	\begin{itemize}
		\item A: healthy people with open eyes (normal 1).
		
		\item B: healthy people with closed eyes (normal 2).
		
		\item C: non-epileptic brain regions that may participate in seizures.
		
		\item D: interictal events.
		
		\item E: ictal events.
	\end{itemize}
	
	Each set contains 100 segments sampled at 173.61 Hz and with a duration of 23.6s each one.
	
	In our experimental evaluations, only the set C was not used due to two reasons: balancing between normal and abnormal segments\footnote{The Epileptologie Bonn database contains 200 normal and 300 abnormal EEG segments.} for binary classifier building and signals of these sets contain patterns similar to interictal segments, which may lead to increased error classification rates~\cite{oliva2018:phd}. In this way, 400 EEG segments were used for feature extraction and classifier building.
	
	For feature extraction, the approaches presented in Section~\ref{sec:features} were computed in each EEG segment, from which a total of 105 features were extracted.

	\subsection{Classifier Building}
	\label{subsec:classifiers}
	
	From the EEG dataset, ML algorithms were applied in two ways: binary and multiclass classification.
	
	For binary classification, the EEG segments used in this work were divided into two balanced groups (normal and abnormal). The normal group is composed of healthy EEG for open and closed eyes. On the other hand, the abnormal group contains interictal and ictal EEG segments. Thus, each group is composed of 200 segments.
	
	For multiclass classification, EEG sets were differentiated into four classes, such as normal 1 (open eyes), normal 2 (closed eyes), interictal, and ictal. The multiclass predictive models were built using the code matrix decomposition method based on AAA approach due to their advantage presented in Section~\ref{subsec:mul_classification}.
	
	For both classification problems, the following widely used ML algorithms were performed:
	
	\begin{itemize}
		\item 1-nearest-neighbor (1NN): performs NN technique by using Euclidean distance for similarity measurement between instances~\cite{alpaydin2014introduction}. 1NN is an implementation of the $k$-nearest-neighbor algorithm for $k = 1$.
		
		\item Backpropagation based on multilayer perceptron (BP-MLP): uses the  backpropagation algorithm to build ANN with structure divided into layers~\cite{haykin2009neural}.
		
		\item J48: implements the C4.5 algorithm~\cite{quinlan2014c4} for DT classifier building.
		
		\item $K$-means based on radial basis function network (KM-RBFN): builds an ANN model using $k$-means clustering algorithm and radial basis function (activation function)~\cite{lowe1988multivariable}.
		
		\item Linear discriminant analysis (LDA): applies LDA method to construct models by linear combination of features, seeking to maximize the separation of training examples by classes~\cite{mclachlan2004discriminant}.
		
		\item Quadratic discriminant analysis (QDA): is a LDA variation which uses quadratic combination of feature~\cite{friedman2001elements}.
		
		\item Random forest (RF): generates a set of DT using the bootstrap aggregating method. In this set, a new example is classified by each DT, in which the most returned class is assigned to the instance~\cite{Breiman2001}.
		
		\item Sequential minimal optimization (SMO): performs SVM method~\cite{chang2011libsvm}. In our experiments, the radial basis function (RBF) and polynomial kernels were used to support the SMO algorithm due to the non-linear characteristics of EEG. Thus, SMO\_RBF and SMO\_Pol classifiers were constructed using RBF and polynomial kernel, respectively.
	\end{itemize}

	\subsection{Performance Evaluation}
	\label{subsec:performance_evaluation}
	
	The classifiers generated by methods presented in Section~\ref{subsec:classifiers} were evaluated by using two approaches: $k$-fold stratified cross-validation (SCV) and confusion matrix (CM).
	
	In the SCV application, the dataset is divided into $k$ equal-sized sets and balanced by class, in which the $i$-th set is used for testing, and the remaining sets are used for training. As a result, $k$ error prediction values are generated and used to calculate performance, such as mean error (ME) and standard deviation (SD)~\cite{mclachlan2005analyzing}. In our experiments, we use 10-fold ($k = 10$) SCV because, besides being the most used approach, its mean error values are close to those obtained when $k$ is equal to the dataset length~\cite{markatou2005analysis}.
	
	Additionally, statistical hypothesis tests for paired were conducted to complement the SCV evaluation by computing error value in order to verify  whether a statistically significant difference among predictive models exists, considering a significance level of 5\%. For choosing an appropriate statistical test, first, it is necessary to verify if the SCV results are within the normal (Gaussian) distribution~\cite{book:fredman:2011}. In this sense, the Shapiro-Wilk normality test~\cite{shapiro1965analysis} was applied due to the small number of error values (10-fold SCV) for each classifier. If the hypothesis test results in a statistical difference, a post hoc test can be performed aiming to identify pairs of predictive models responsible for this result~\cite{book:fredman:2011}.
	
	In another evaluation approach, the CM was used to measure the classifier performance for each class. This matrix is commonly used to evaluate binary classifiers, \textit{i.e}, it is composed by four elements (Table~\ref{tab:cm})~\cite{fawcett2006introduction,book:fredman:2011}:
	
	\begin{itemize}
		\item True positive (TP): number of abnormal (positive) examples classified as abnormal.
		
		\item False negative (FN): number of abnormal examples classified as normal (negative).
		
		\item True negative (TN): number of normal examples classified as normal.
		
		\item False positive (FP): number of normal examples classified as abnormal.
	\end{itemize}

	\begin{table}[!htp]
		\centering
		\renewcommand{\arraystretch}{1.10}
		\scalefont{0.85}
		\caption{Confusion matrix example for binary classifiers.}
		\label{tab:cm}
		\begin{tabular}{ccc}
			& \textbf{positive} & \textbf{negative} \\ \cline{2-3} 
			\multicolumn{1}{c|}{\textbf{positive}} & \multicolumn{1}{c|}{TP} & \multicolumn{1}{c|}{FP} \\ \cline{2-3} 
			\multicolumn{1}{c|}{\textbf{negative}} & \multicolumn{1}{c|}{FN} & \multicolumn{1}{c|}{TN} \\ \cline{2-3} 
		\end{tabular}
	\end{table}

	As presented in Table~\ref{tab:cm}, in a CM, the $i$-th row corresponds to examples predicted as $i$-th class and the $j$-th column presents predictions attributed to examples from the $j$-th class. In this way, several measures can be extracted from a CM, such as:
	
	\begin{itemize}
		\item Sensitivity or specificity: rate of examples for a given class (positive or negative) correctly classified. For instance, sensitivity (TP rate) can be obtained using Equation~\ref{eq:sen}.
		
		\begin{equation}
		\label{eq:sen}
		TP\text{ }rate = \frac{TP}{TP + FN}
		\end{equation}

		\item Predictive value (PV): from all examples classified as a particular class, PV is the rate of those correctly classified. For example, positive PV can be obtained using Equation~\ref{eq:ppv}.
		
		\begin{equation}
		\label{eq:ppv}
		Positive\text{ }PV = \frac{TP}{TP + FP}
		\end{equation}
	\end{itemize}

	Since CM and its measures are related to classification performance by class, they can also be applied to evaluate multiclass predictive models. Thereby, in a CM, the element $[i, j]$ is the amount of $i$ examples classified as $j$. For example, the diagonal elements of a CM refer to correctly classified examples. Finally, from a multiclass CM, correctly classified rate (CCR) and PV for the $i$-th class can be obtained from Equation \ref{eq:ccr} and \ref{eq:pv}, respectively, where $n$ is the number of classes.
	
	\begin{equation}
	\label{eq:ccr}
	CCR(i) = \frac{CM_{i,i}}{\sum\limits_{j = 1}^{n}CM_{i,j}}
	\end{equation}
	
	\begin{equation}
	\label{eq:pv}
	PV(i) = \frac{CM_{i,i}}{\sum\limits_{j = 1}^{n}CM_{j,i}}
	\end{equation}

	\section{Results and Discussions}
	\label{sec:results}

	\subsection{Stratified cross-validation}
	\label{subsec:cross_validation_res}
	
	Table \ref{tab:SCV} presents the SCV results achieved by binary and multiclass classifiers.
	
	\begin{table}[ht!]
		\centering
		\renewcommand{\arraystretch}{1.10}
		\caption{SCV results for binary and multiclass classifiers.}
		\label{tab:SCV}
		\begin{tabular}{|l|c|c|c|c|}
			\hline
			\multirow{2}{*}{\textbf{Algorithm}} & \multicolumn{2}{c|}{\textbf{Binary}} & \multicolumn{2}{c|}{\textbf{Multiclass}} \\ \cline{2-5} 
			& \textbf{ME (\%)}  & \textbf{SD (\%)} & \textbf{ME (\%)}    & \textbf{SD (\%)}   \\ \hline
			\textbf{LDA}      & 2.75          & 3.22          & 6.25          & 4.45          \\ \hline
			\textbf{BP-MLP}   & 2.00          & 2.58          & \textbf{3.75} & 3.17          \\ \hline
			\textbf{QDA}      & 1.75          & \textbf{1.69} & 7.25          & 4.48          \\ \hline
			\textbf{KM-RBFN}  & 4.75          & 5.71          & 11.00         & 5.16          \\ \hline
			\textbf{SMO\_Pol} & 3.75          & 3.58          & 4.50          & \textbf{2.58} \\ \hline
			\textbf{SMO\_RBF} & 1.75          & 2.37          & 4.50          & 3.69          \\ \hline
			\textbf{1NN}      & 2.50          & 3.12          & 6.25          & 3.58          \\ \hline
			\textbf{RF}       & \textbf{1.25} & 1.77          & 4.75          & 4.48          \\ \hline
			\textbf{J48}      & 3.00          & 2.30          & 11.25         & 4.45          \\ \hline
		\end{tabular}
	\end{table}

	According to Table~\ref{tab:SCV}, the RF predictive model classified the highest amount of EEG correctly, for which a 98.75\% accuracy rate was recorded, \textit{i.e.}, only five (two normal and three abnormal) segments were incorrectly labeled. The QDA algorithm trained the most stable binary model, for which such fact was evidenced by the lowest error rate dispersion (SD of 1.69\%) obtained during the SCV application. 
	
	Features based on SG and BG complemented PS measures for the building of the most accurate binary RF model, since its respective algorithm uses importance measures, generate during bootstrap aggregation method application~\cite{ho1995random}, for attribute selection, which allowed the choice of the most representative features. In contrast, the binary KM-RBFN model obtained the lowest accuracy (95.25\%) and found the highest dispersion of classification error values (SD of 5.71\%) for EEG differentiation.
	
	Regarding the multiclass problem, the BP-MLP classifier achieved the lowest ME value, which was 3.75\%, \textit{i.e.}, 96.25\% accuracy. The SMO algorithm trained the most stable multiclass model using the polynomial kernel (SMO\_Pol), for which such fact was evidenced by the lowest error rate dispersion (SD of 2.58\%) obtained during the SCV application. The high hit rate of model BP-MLP was due to high generalization of nonlinear patterns by the respective algorithm.
	
	Although the multiclass J48 classifier has achieved the worst accuracy (88.75\%), the KM-RBFN model was the most unstable, for which a 5.16\% SD was obtained. The main reason for classification errors was in the differentiation between normal (open and closed eyes) EEG, whose extracted features can result in similar patterns for some cases, for example, changes in alpha waves (more evident in closed-eye EEG) into open-eye segments.
	
	In both classification problems, all built classifiers are accurate, since several extracted features (\textit{e.g.} PS average) contain linearly separable patterns in some cases, for instance, ictal segments tend to generate measurements with higher values in relation to the signals of other classes.
	
	In addition to the experimental evaluations by using SCV (Table~\ref{tab:SCV}), whose results are presented in Section \ref{subsec:cross_validation_res}, a statistical hypothesis tests for paired data were conducted considering a significance level of 5\%. These tests were computed in order to check if there is a statistical difference among predictive models in each classifier case. Prior to this, the Shapiro-Wilk normality test~\cite{shapiro1965analysis} was applied to define the appropriate test (parametric or non-parametric) for this study. The normality test results are presented in Table~\ref{tab:shap}.

	\begin{table}[ht!]
		\renewcommand{\arraystretch}{1.10}
		\centering
		\caption{Shapiro-Wilk test results.}
		\label{tab:shap}
		\begin{tabular}{|l|c|c|}
			\hline
			\multicolumn{1}{|c|}{\multirow{2}{*}{\textbf{Algorithm}}} & \multicolumn{2}{c|}{\textbf{\textit{P}-value}} \\ \cline{2-3} 
			\multicolumn{1}{|c|}{} & \textbf{Binary} & \textbf{Multiclass} \\ \hline
			\textbf{LDA}      & 0.0284 & 0.3354 \\ \hline 
			\textbf{BP-MLP}   & 0.0114 & 0.2375 \\ \hline 
			\textbf{QDA}      & 0.0154 & 0.2462 \\ \hline 
			\textbf{KM-RBFN}  & 0.0001 & 0.4473 \\ \hline 
			\textbf{SMO\_Pol} & 0.1511 & 0.0114 \\ \hline 
			\textbf{SMO\_RBF} & 0.0006 & 0.2247 \\ \hline
			\textbf{1NN}      & 0.0068 & 0.1511 \\ \hline
			\textbf{RF}       & 0.0021 & 0.0549 \\ \hline
			\textbf{J48}      & 0.0042 & 0.6327 \\ \hline
		\end{tabular}
	\end{table}

	According to Table~\ref{tab:shap}, only the SMO\_Pol binary classifier passed in the normality test, \textit{i.e.} it reached p-values $> 0.05$, noting that its SCV results are within the Gaussian distribution. For the multiclass case, only SMO\_Pol predictive model failed the normality test.
	
	Thus, since the normality test evidenced that there were predictive models out of Gaussian distribution for both classification cases, a non-parametric statistical hypothesis test must be performed. For this, the Friedman test~\cite{book:fredman:2011} was used.
	
	The Friedman test concerning binary classifiers resulted in a $p$-value of 0.1398, stating that a statistical difference was not among predictive models. For multiclass classifiers, a $p$-value of 0.0002, evidencing highly significant statistical difference among predictive models.
	
	For the identification of the pairs of multiclass classifiers statistically different, the Nemenyi post hoc test~\cite{nemenyi1962distribution} was applied. Table \ref{tab:pht_mul} presents the comparisons in which statistical difference was detected for multiclass classifiers.

	\begin{table}[ht!]
		\renewcommand{\arraystretch}{1.10}
		\centering
		\caption{Main post-hoc test results for multiclass classifiers.}
		\label{tab:pht_mul}
		\begin{tabular}{|c|c|}
			\hline
			\textbf{Pair of classifiers} & \textbf{\textit{P}-value} \\ \hline
			BP-MLP \textit{vs.} KM-RBFN   & 0.0073 \\ \hline
			BP-MLP \textit{vs.} J48       & 0.0202 \\ \hline
			SMO\_RBF \textit{vs.} KM-RBFN & 0.0499 \\ \hline
		\end{tabular}
	\end{table}

	As shown in Table~\ref{tab:pht_mul}, it can only be stated, with 95\% certainty, the KM-RBFN classifier obtained inferior performance compared to the BP-MLP and SMO\_RBF models. Another comparison proved the superiority of the BP-MLP model over the J48.
	
	Finally, although it was not possible to statistically define the best ML algorithm for epilepsy detection in this study, both experimental evaluations (binary and multiclass classification) reached promising results. In contrast, statistical tests proved that KM-RBFN and J48 models had lower performance in comparison to some other multiclass classifiers.

	\subsection{Confusion matrix measures}
	
	In Table~\ref{tab:bin_cm}, CCR and PV results obtained from binary classifiers are presented.
	
	\begin{table}[ht!]
		\renewcommand{\arraystretch}{1.10}
		\centering
		\caption{CCR and PV results for binary classifiers.}
		\begin{tabular}{|l||c|c||c|c|}
			\hline
			& \multicolumn{2}{c||}{\textbf{CCR} (\%)} & \multicolumn{2}{c|}{\textbf{PV} (\%)} \\ \hline
			\multicolumn{1}{|c||}{\textbf{Algorithm}} & \textbf{Normal} & \textbf{Abnormal} & \textbf{Normal} & \textbf{Abnormal} \\ \hline
			\textbf{LDA}       & 97.50           & 97.00          & 97.01          & 97.49           \\ \hline
			\textbf{BP-MLP}    & 98.00           & 98.00          & 98.00          & 98.00           \\ \hline
			\textbf{QDA}       & 99.00           & 97.50          & 97.54          & 98.98           \\ \hline
			\textbf{KM-RBFN}    & \textbf{100.00} & 90.50          & 91.32          & \textbf{100.00} \\ \hline
			\textbf{SMO\_Pol}  & 96.50           & 96.00          & 96.02          & 96.48           \\ \hline
			\textbf{SMO\_RBF} & 97.50           & \textbf{99.00} & \textbf{98.98} & 97.54           \\ \hline
			\textbf{1NN}       & 96.50           & 98.50          & 98.47          & 96.57           \\ \hline
			\textbf{RF}        & 99.00           & 98.50          & 98.51          & 98.99           \\ \hline
			\textbf{J48}       & 96.50           & 97.50          & 97.47          & 96.53           \\ \hline
		\end{tabular}
		\label{tab:bin_cm}
	\end{table}

	As presented in Table~\ref{tab:bin_cm}, the KM-RBFN model correctly classified the highest amount of normal EEG segments (CCR of 100\%) and was more likely to detect abnormal signals (PV of 100\%) in comparison to other classifiers. On the other hand, SMO\_RBF classifier correctly predicted the highest amount of abnormal EEG (CCR of 99\%) and was more likely to detect normal segments (PV of 98.98\%) in comparison to other classifiers.
	
	CCR and PV resulted from multiclass predictive models are presented in Table \ref{tab:mul_CCR} and \ref{tab:mul_PV}, respectively.
	
	\begin{table}[ht!]
		\renewcommand{\arraystretch}{1.10}
		\centering
		\caption{CCR results for multiclass classifiers.}
		\begin{tabular}{|l|c|c|c|c|}
			\hline
			\multicolumn{1}{|c|}{\textbf{Algorithm}} & \textbf{\begin{tabular}[c]{@{}c@{}}Normal 1\\ (\%)\end{tabular}} & \textbf{\begin{tabular}[c]{@{}c@{}}Normal 2\\ (\%)\end{tabular}} & \textbf{\begin{tabular}[c]{@{}c@{}}Interictal\\ (\%)\end{tabular}} & \textbf{Ictal (\%)} \\ \hline
			\textbf{LDA}      & \textbf{100.00} & 90.00          & 96.00          & 89.00           \\ \hline
			\textbf{BP-MLP}   & 95.00           & 95.00          & \textbf{97.00} & 98.00           \\ \hline
			\textbf{QDA}      & 89.00           & \textbf{98.00} & 84.00          & \textbf{100.00} \\ \hline
			\textbf{KM-RBFN}  & 97.00           & 87.00          & 87.00          & 85.00           \\ \hline
			\textbf{SMO\_Pol} & 95.00           & 93.00          & 96.00          & 98.00           \\ \hline
			\textbf{SMO\_RBF} & 91.00           & 96.00          & 96.00          & 99.00           \\ \hline
			\textbf{1NN}      & 89.00           & 94.00          & 96.00          & 96.00           \\ \hline
			\textbf{RF}       & 96.00           & 94.00          & 95.00          & 96.00           \\ \hline
			\textbf{J48}      & 89.00           & 87.00          & 85.00          & 94.00           \\ \hline
		\end{tabular}
		\label{tab:mul_CCR}
	\end{table}

	\begin{table}[htp!]
		\renewcommand{\arraystretch}{1.10}
		\centering
		\caption{PV results for multiclass classifiers.}
		\begin{tabular}{|l|c|c|c|c|}
			\hline
			\multicolumn{1}{|c|}{\textbf{Algorithm}} & \textbf{\begin{tabular}[c]{@{}c@{}}Normal 1\\ (\%)\end{tabular}} & \textbf{\begin{tabular}[c]{@{}c@{}}Normal 2\\ (\%)\end{tabular}} & \textbf{\begin{tabular}[c]{@{}c@{}}Interictal\\ (\%)\end{tabular}} & \textbf{Ictal (\%)} \\ \hline
			\textbf{LDA}      & 86.96 & 96.77 & 93.20 & \textbf{100.00} \\ \hline
			\textbf{BP-MLP}   & 92.23 & 95.96 & 97.00 & \textbf{100.00} \\ \hline
			\textbf{QDA}      & \textbf{98.89} & 89.09 & \textbf{98.82} & 86.96  \\ \hline
			\textbf{KM-RBFN}  & 80.17 & 93.55 & 87.88 & 97.70  \\ \hline
			\textbf{SMO\_Pol} & 91.35 & 96.88 & 96.00 & 98.00  \\ \hline
			\textbf{SMO\_RBF} & 94.79 & 93.20 & 97.96 & 96.12  \\ \hline
			\textbf{1NN}      & 89.00 & 94.00 & 93.20 & 98.97  \\ \hline
			\textbf{RF}       & 94.12 & \textbf{96.91} & 94.06 & 96.00  \\ \hline
			\textbf{J48}      & 83.18 & 89.69 & 89.47 & 93.07  \\ \hline
		\end{tabular}
		\label{tab:mul_PV}
	\end{table}

	According to Table~\ref{tab:mul_CCR}, LDA and QDA predictive models correctly classified all EEG segments from normal1 (closed-eye healthy) and ictal classes, respectively. Additionally, QDA was also the most accurate for classification of normal closed-eye EEG, for which 98\% CCR was reached. For the prediction of interictal segments, the BP-MLP achieved the highest rating rate, with 97\% CCR.
	
	As highlighted in Table~\ref{tab:mul_PV}, the QDA and RF classifiers are most likely to correctly predict normal segments, for which PV values of 98.89\% (normal 1) and 96.91\% (normal 2) were obtained, respectively. Interictal EEG is more likely to be correctly classified by the QDA, whose PV was 98.82\%. Lastly, ictal segments are more likely to be predicted by LDA and BP-MLP classifiers, which reached a PV of 100\%.
	
	As with evaluation based on SCV, through the CM it was not possible to identify a classifier that actually stands out in relation to other predictive models for EEG classification. Nevertheless, all classifiers achieved competitive performance.

	\subsection{Comparisons with related works}
	
	Concerning related works, our most accurate classifiers reached competitive performances. For binary classification, the RF model correctly classified 98.75\% EEG segments. For multiclass classification, BP-MLP achieved 96.25\% accuracy to differentiate normal 1, normal 2, interictal, and ictal EEG segments. Also, CM measures, such as CCR and PV, were used in the evaluation aiming to obtain performance estimates by class. In binary classification, CCR of abnormal and normal classes are equivalent to sensitivity and specificity, respectively. Another binary equivalence is about the PV measure, which can be positive (abnormal) or negative (normal) PV.
	
	Table~\ref{tab:bin_comp} shows a comparison among the best performance values reached by binary classifiers. In this table, only \cite{wang2018epileptic} and \cite{wu2019intelligent} were not included because a different EEG database was used in their respective experiments.
	
	\begin{table}[htp!]
		\renewcommand{\arraystretch}{1.10}
		\centering
		\caption{Performance comparison among accurate binary classifiers.}
		\begin{tabular}{|c|c|c|c|c|c|}
			\hline
			\multirow{2}{*}{\textbf{Work}} & \multicolumn{2}{c|}{\textbf{CCR (\%)}} & \multicolumn{2}{c|}{\textbf{PV (\%)}} & \multirow{2}{*}{\textbf{Accuracy (\%)}} \\ \cline{2-5}
			& \textbf{Abnormal} & \textbf{Normal} & \textbf{Abnormal} & \textbf{Normal} &  \\ \hline
			\cite{jaiswal2017local}    & ---    & ---    & ---    & ---   & 99.90 \\ \hline
			
			\cite{yavuz2018epileptic}  & 100.00 & 100.00 & ---    & ---   & 100.00 \\ \hline
			
			\cite{goksu2018eeg}        & 100.00 & 100.00 & ---    & ---   & 100.00 \\ \hline
			
			\cite{subasi2019epileptic} & 100.00 & 99.50  & ---    & ---   & 99.38 \\ \hline
			
			\cite{sayeed2019neuro}     & 100.00 & 100.00 & 100.00 & ---   & 100.00 \\ \hline
			
			\cite{gao2020automatic}    & 99.17  & 92.75  & ---    & ---    & 92.00 \\ \hline
			
			Ours                        & 100.00 & 99.0   & 98.98  & 100.00 & 98.75 \\ \hline
		\end{tabular}
		\label{tab:bin_comp}
	\end{table}
	
	As exhibited in Table~\ref{tab:bin_comp}, our most accurate model (RF) achieved the highest evaluation measures only compared to \cite{gao2020automatic}. Even so, the RF had a performance considered competitive for differentiation among normal and abnormal, since it failed to predict only five segments (from 400). In related works, EEG differentiation was normal \textit{vs.} ictal or/and interictal\textit{ vs.} ictal, being that in these comparisons, the patterns are commonly linearly separable by class in the feature space, reducing the complexity for classifier building. In our study, two EEG groups were used to build binary classifiers: normal (open and closed eyes) and abnormal (interictal and ictal). In other words, we used 400 EEG segments to generate models, while 200 segments were used to build the most accurate classifiers in the related works presented in Table~\ref{tab:bin_comp}. In a previous study~\cite{oliva2017epileptic}, we also got to achieve 100\% accuracy for differentiation between normal and ictal signals.
	
	Table~\ref{tab:mul_comp} presents a comparison among the best performance values achieved by multiclass predictive. In this comparison, only \cite{shen2017ga} was not considered due to the same reason presented in Table~\ref{tab:bin_comp}.
	
	\begin{table}
		\centering
		\caption{Performance comparison among accurate multiclass classifiers.}
		\begin{tabular}{|c|c|}
			\hline
			\textbf{Work} & \textbf{Accuracy (\%)} \\ \hline
			\cite{zeng2016automatic}     & 76.70 \\ \hline
			
			\cite{jaiswal2017local}      & 99.90 \\ \hline
			
			\cite{sikdar2018epilepsy}    & 99.60 \\ \hline
			
			\cite{yavuz2018epileptic}    & 98.85 \\ \hline
			
			\cite{goksu2018eeg}          & 100.00 \\ \hline
			
			\cite{ren2019classification} & 99.43 \\ \hline
			
			Ours                         & 96.25 \\ \hline
		\end{tabular}
		\label{tab:mul_comp}
	\end{table}

	As shown in Table~\ref{tab:mul_comp}, our most accurate classifier (BP-MLP) also outperformed only a related work for the measure considered (accuracy) in the comparisons. Our predictive models were built to differentiate EEG segments among four different classes (normal 1 \textit{vs.} normal 2 \textit{vs.} interictal \textit{vs.} ictal), while in related works presented in Table~\ref{tab:mul_comp}, only three classes were considered. One of the main causes of classification errors in our work was the difficulty in differentiating between normal 1 and 2 EEG segments, whose patterns could be similar (as presented in Section~\ref{subsec:cross_validation_res}). Nevertheless, our classifiers achieved competitive results. Also, unlike related works, we calculate predictive measures (CCR and PV) for each class for multiclass predictive.

	\section{Conclusion}
	\label{sec:conclusion}
	
	Several feature extraction and machine learning (ML) methods were applied in this work in order to build binary and multiclass predictive models for epilepsy detection, for which open eyes normal, closed eyes normal, interictal, and ictal EEG segments were differentiated. Multitaper was used to generate power spectrum, spectrogram, and bispectrogram, from which, a total of 105 measures were computed. Binary and multiclass predictive models were built using 1-nearest-neighbor (1NN), backpropagation multilayer perceptron (BP-MLP), linear discriminant analysis (LDA), quadratic discriminant analysis (QDA), J48, random forest (RF), $K$-means based on radial basis function network (KM-RBFN), and sequential minimal optimization (SMO) algorithms, the latter being supported by nonlinear and radial basis function kernels, were used to build binary and multiclass predictive models.
	
	For binary classification, 400 EEG segments were differentiated between normal (healthy open and closed eye) and abnormal (interictal and ictal) classes. As a result, classifiers reached accuracy between 95.25\% and 98.75\%. The Friedman test, considering the significance level of 5\%, did not find a significant statistical difference.
	
	For multiclass classification, the code matrix decomposition method based on all-against-all approach was used to generate classifiers able to differentiate EEG segments among normal 1 (open eye), normal 2 (closed eye), interictal, and ictal classes. In the evaluation, the predictive models obtained accuracy between 88.75\% and 96.25\%. As in the previous classification case, it was not possible to find a model that achieved better performance through statistical tests and measures based on confusion matrices.
	
	Nevertheless, the binary and multiclass classifiers achieved competitive results with the literature. Also, our results can be improve by using other features (\textit{e.g.} fractal dimension, line length, etc), attribute selection techniques, and parameter tuning of ML algorithms.
	
	As much as methods based on deep learning have gained attention in recent years, convention ML methods are still a good tool for solution of classification problems, since they are computationally more economical for not very large databases and the resulting models can be less complex generalizations to be understood. Also, for the training of accurate classifiers, it is not the use of large databases (commonly redundant) for generalization of classification problems, but the choice of representative examples and appropriate measures (features), although it may be considered difficult in several cases.

%
%
%
%
%

	\section*{Acknowledgments}
	
	J. T. Oliva would like to thank the Brazilian funding agency Coordena\c{c}{\~a}o de Aperfei\c{c}oamento de Pessoal de N{\'i}vel Superior (CAPES) for financial support. J. L. G. Rosa is grateful to the Brazilian agency S{\~a}o Paulo Research Foundation (FAPESP -- process 2016/02555-8) for the financial support.


\end{document}